\documentclass[conference]{IEEEtran}
\IEEEoverridecommandlockouts
\usepackage[T1]{fontenc}
\usepackage{graphics} 
\usepackage[pdftex]{graphicx} 
\usepackage{amsmath} 
\usepackage{amssymb}  
\usepackage{todonotes}
\usepackage{acro}
\usepackage{balance}
\usepackage{subcaption}
\usepackage{siunitx}
\usepackage[noadjust]{cite}
\usepackage{hyperref}
\usepackage[font={footnotesize}]{caption}
\usepackage{verbatim}
\usepackage{adjustbox}

\graphicspath{{chapters/images/}}

\DeclareAcronym{slam}{short = SLAM, long = Simultaneous Localization and Mapping}
\DeclareAcronym{cnn}{short = CNN, long = Convolutional Neural Network}
\DeclareAcronym{vi}{short = VI, long = visual-inertial}

\newcommand{\modelsynth}{$\text{m}_{\text{synth}}$}
\newcommand{\modelreal}{$\text{m}_{\text{real}}$}
\newcommand{\modelmix}{$\text{m}_{\text{mix}}$}
\newcommand{\fone}{$\text{F}_1$ }

\def\BibTeX{{\rm B\kern-.05em{\sc i\kern-.025em b}\kern-.08em
    T\kern-.1667em\lower.7ex\hbox{E}\kern-.125emX}}

\title{Dynamic Objects Segmentation for Visual Localization in Urban Environments
\author{G. Zhou$^1$, B. Bescos$^2$, M. Dymczyk$^1$, M. Pfeiffer$^1$, J. Neira$^2$, R. Siegwart$^1$}
\thanks{This work has received funding from the European Union (Horizon 2020), project CROWDBOT Grant No. 779942, and the Spanish Ministry of Economy and Competitiveness (project DPI2015-68905-P)} 
\thanks{$^1$ Autonomous Systems Lab, ETH Zurich; \, $^2$  Instituto de Investigaci\'on en Ingenier\'ia de Arag\'on, Universidad de Zaragoza }
}

\begin{document}
\bstctlcite{IEEEexample:BSTcontrol}
\maketitle

\begin{abstract}
Visual localization and mapping is a crucial capability to address many challenges in mobile robotics. 
It constitutes a robust, accurate and cost-effective approach for local and global pose estimation within prior maps.
Yet, in highly dynamic environments, like crowded city streets, problems arise as major parts of the image can be covered by dynamic objects.
Consequently, visual odometry pipelines often diverge and the localization systems malfunction as detected features are not consistent with the precomputed 3D model.

In this work, we present an approach to automatically detect dynamic object instances to improve the robustness of vision-based localization and mapping in crowded environments.
By training a convolutional neural network model with a combination of synthetic and real-world data, dynamic object instance masks are learned in a semi-supervised way. 
The real-world data can be collected with a standard camera and requires minimal further post-processing.
Our experiments show that a wide range of dynamic objects can be reliably detected using the presented method.
Promising performance is demonstrated on our own and also publicly available datasets, which also shows the generalization capabilities of this approach.
\end{abstract}

\section{Introduction}
\label{sec:introduction}
While robot navigation in static environments is nowadays well understood, there are still many open research questions when leaving controlled research environments and starting to navigate in the real world.
In highly dynamic environments, like dense pedestrian crowds, new challenges arise which require specific tools for perception, localization and path planning. 
Regarding motion planning in those environments, a significant amount of work has been presented within the last years, covered in \cite{pfeiffer2016iros,pfeiffer2017lstm,kretzschmar2016social,vemula2017modeling,pettre2009experiment}, to name a few. 
However, for robot navigation, robust localization systems are of similar importance. 
Vision-based \ac{slam} offers a cost effective and reliable localization solution \cite{schneider2018maplab, lynen2015get, mur2015orb} for a broad range of mobile robotics. 

Both for map building and localization in a prior map, visual \ac{slam} systems rely on a combination of feature detection and tracking. 
The former approach yields several problems in highly dynamic environments: 
As a major part of the image can be covered by moving objects, a significant amount of features will be detected on them, which leads to subpar quality of odometry estimation, map building and global localization. 
Furthermore, the maps contain numerous non-persistent 3D landmarks that increase the map size and introduce inconsistencies, \textit{e.g.}, if one vehicle is observed in different spots. 

In this work, we present a novel approach to detect dynamic image regions using a \ac{cnn} based approach. 
This approach will provide dynamic object instance masks to vision-based localization frameworks, like \textit{maplab} \cite{schneider2018maplab} or \textit{ORB-SLAM} \cite{mur2015orb}, on a frame-by-frame basis.
By using this information, the \ac{slam} frameworks can ideally only rely on persistent and static information and exclude any dynamic areas of the image.
Therefore, the robustness and precision of such frameworks in dense crowds can be significantly improved and map size can be reduced.

\begin{figure}[t]
    \centering
    \includegraphics[width=\linewidth]{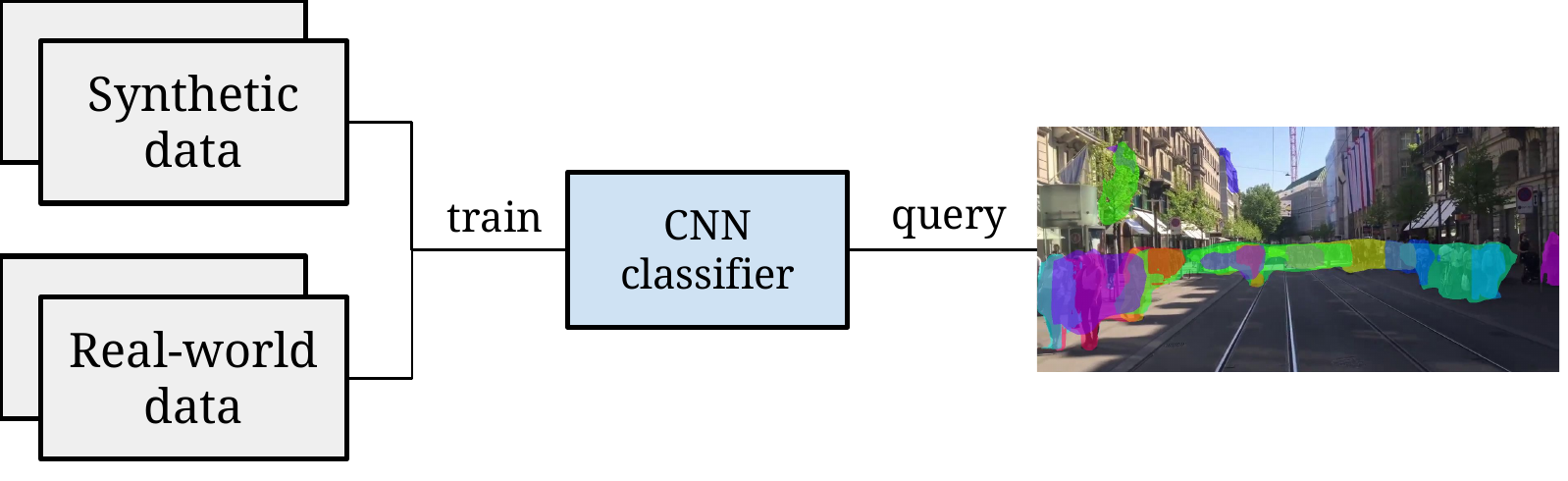}
    \caption{The presented CNN classifier can extract dynamic objects instances masks from a single image. 
    It is trained in a self-supervised fashion with a combination of synthetic and real-world data, collected using a RGB camera.}
    \label{fig:teaser}
    \vspace{-5mm}
\end{figure}

The presented approach builds upon the \textit{Mask R-CNN} model~\cite{he2017mask}, which allows for instance based image semantic segmentation. 
It learns to segment dynamic parts of the image in a semi-supervised way, without being restricted to certain semantic object classes. 
We use a combination of synthetic data, which has been generated with the driving simulator CARLA~\cite{dosovitskiy2017carla}, and real-world video snippets to train the instance segmentation model. 
We hypothesize that our approach will be able to learn general dynamic objects, even if no specific semantic labels are present for training (Figure~\ref{fig:teaser}).

The main contributions of this work are: 
(i) An approach for the unsupervised mask extraction of dynamic instances in real-world images, 
(ii) a learning-based model for frame-by-frame dynamic instance segmentation and 
(iii) the evaluation on a popular open-source dataset.


\section{Related Work}
\label{sec:related_work}
Most \ac{slam} systems classify non-static objects as spurious data and reject features.
Common outlier rejection algorithms are RANSAC (\textit{e.g.}, in ORB-SLAM \cite{mur2015orb}) and robust cost functions (\textit{e.g.}, in PTAM \cite{klein2007parallel}).
There are several SLAM systems that more specifically address the dynamic scene content, trying to separate dynamic and static regions within the images.
Alcantarilla~\textit{et~al.}~\cite{alcantarilla2012combining} detect moving objects by means of a scene flow representation with stereo cameras. 
Tan \textit{et~al.}~\cite{tan2013robust} detect changes that take place in the scene by projecting the map features into the current frame for appearance and structure validation. 
By using \mbox{RGB-D} cameras, Kim~\textit{et~al.}~\cite{kim2016effective} propose to obtain the static parts of the scene by computing the difference between consecutive depth images projected over the same plane. 
More recently, the work of Li and Lee~\cite{li2017rgb} uses depth edges points for odometry estimation, which have an associated weight indicating their probability of belonging to a dynamic object.
Sun \textit{et~al.}~\cite{sun2017improving} calculate the difference in intensity between consecutive RGB images for further pixel classification within the quantized depth image.

While these methods succeed in detecting motion, they lack semantic understanding of the scene.
As a result, only objects moving at the time of data collection can be detected, and dynamic objects which remain temporarily static would not be correctly identified, becoming part of the 3D map, eventually leading to failure in long-term SLAM scenarios. 

A standard approach to detect the known dynamic content in the scene is the use of a \ac{cnn} for pixel-wise semantic segmentation, like~\cite{he2017mask,badrinarayanan2015segnet}. 
Such systems succeed in detecting objects that are known to move, \textit{i.e.}, people, vehicles, \textit{etc}., but fail to detect new dynamic classes.
By contrast, Guizilini and Ramos~\cite{guizilini2015online} apply a two-level classification mechanism (RANSAC, Gaussian Process) to detect and further learn the dynamic regions in images.
Recently, the work of Barnes \textit{et~al.}~\cite{barnes2017driven} has shown how to train a pixel-level segmentation classifier to identify dynamic and static regions in the images. 
They leverage large-scale offline mapping approaches with LiDAR to obtain ephemerality masks for images. 
Other works also use machine learning~\cite{hartmann2014predicting} and deep learning~\cite{dymczyk2016will, bescos2018dynslam} to deal with dynamic objects.
These approaches are capable of detecting, on a single-view basis, yet require a specific and expensive sensor suite, a significant amount of calculations and multiple mapping sessions.

In contrast to those approaches, we propose a system that relies on single RGB frames to identify dynamic objects and train a classifier to reject spurious data. 

\section{Approach}
\label{sec:approach}

The contribution of this paper is an approach to detect dynamic object instances in an image on a single frame basis.
The method relies on \acp{cnn}, as they demonstrated an outstanding performance in image analysis and understanding over the recent years~\cite{girshick2015fast,he2017mask,badrinarayanan2015segnet}. 
This section consists of two major parts:
In~\ref{subsec:model}, we introduce a model suited to detect dynamic instances in images.
In the succeeding Section~\ref{subsec:train_data}, we present the training data preparation, which is able to extract dynamic objects from real-world data.

\subsection{Neural network model}
\label{subsec:model}
We adopt \textit{Mask R-CNN}~\cite{he2017mask}, which is specifically tailored to instance detection. 
The model has two parts: the convolutional \emph{backbone}, which is mostly responsible for feature extraction, and the \emph{head} that computes the region proposals and segmentation masks. 
We resort to the pre-trained backbone weights\footnote{\href{https://github.com/matterport/Mask_RCNN}{\tt https://github.com/matterport/Mask\_RCNN}} based on the COCO dataset~\cite{lin2014microsoft} and only fine-tune the head module using our own data.
The head module is modified for binary classification between the classes static and dynamic. 

\subsection{Training data generation}
\label{subsec:train_data}

\begin{figure*}[htbp]
    \centering
    \includegraphics[width=\linewidth]{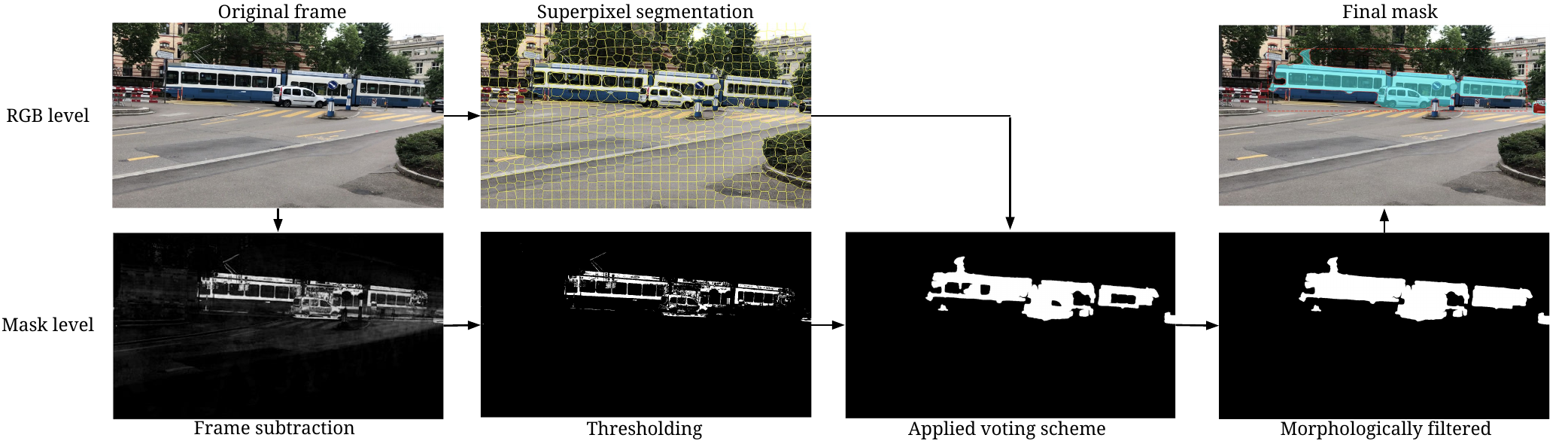}
    \caption{The pipeline for image mask extraction from a set of several frames.
    The moving parts of a frame are found by subtraction of other frames from the same location, followed by thresholding, a voting scheme and super-pixel filtering. 
    Finally, a morphological filter is applied for further refinement until the final image mask is obtained.
    The final mask accurately covers the dynamic object with a single component.}
    \label{fig:image_filtering}
    \vspace{-5mm}
\end{figure*}

To train a binary classifier, binary image masks need to be provided (see Figure~\ref{fig:teaser}). 
We train the model using both synthetic and real-world data.
Most of the training samples are sourced from the CARLA driving simulator~\cite{dosovitskiy2017carla}.
Then, we augment them with real-world samples that need to be manually collected.
The synthetic data can be easily generated and also annotated using the semantic segmentation provided by the rendering engine.
In order to avoid manual labeling of the real-world data, we propose an approach to obtain image masks from short video clips recorded with a static RGB camera, without any human intervention.
The approach consists of 5 steps as shown in Figure~\ref{fig:image_filtering}:
\paragraph{Frame subtraction}
\label{appr:diff_frame}
Several short video clips are recorded in dynamic urban environments using a static camera.
Each clip is taken at another location.
Single frames are extracted from each video clip to provide an original frame set (OFS). 
The training frame set (TFS) is then obtained by sampling 5 images from each OFS.

For each query image in the TFS, all images from the OFS (except the query image) are subtracted from the query image to get absolute difference frames (ADFs) -- high pixel value means a large difference.
Given that the camera is static, pixel differences correspond to changes in the environment which are caused by moving objects.

\paragraph{Thresholding}
To obtain binary instances masks, a threshold is applied to the ADFs to form binary difference frames (BDFs).
The threshold value is chosen to be $\mu_p + \sigma_p$, where $\mu_p$ represents the mean and $\sigma_p$ the standard deviation of the pixels intensity in each ADF.
In BDFs, the value 1 represents a dynamic object and 0 the static background.

\paragraph{Voting scheme}

All elements in BDF set are summed up into one single image. 
Large intensity pixels belong to dynamic regions in the query frame, as those will differ from all the other OFS images.
More specifically, those pixels whose intensity is above the threshold value $\tau_{c} \times cardinality$~(BDF) are marked as dynamic. 
The value of $\tau_{c}$ is set to $0.65$ as a good trade-off between true and false positives. 

\paragraph{Super-pixel segmentation}
The result of the previous pixel-wise operations are sparse and noisy masks as shown in Figure~\ref{fig:image_filtering}.
To address this, we partition the image into \textit{super-pixels}~\cite{ren2003learning}. 
If the percentage of dynamic pixels within a single super-pixel exceeds $5\%$, the entire super-pixel is labeled as dynamic and holes inside dynamic masks are filled.

\paragraph{Morphological filtering}
At this stage, parts of a~single dynamic object can be disjoint (see the tram in Figure~\ref{fig:image_filtering}).
To reduce this effect, dilation with a size of $5 \times 5$ pixels is applied to each connected component. 
Afterwards, only the connected components with enough dynamic pixels are kept and subsequently eroded to undo the shape change caused by dilation. 
The final segmentation result for the training data is presented in Figure \ref{fig:image_filtering} (top-right).


\section{Experiments}
\label{sec:experiments}
In this section the performance of the presented approach is analyzed both quantitatively and qualitatively. 
As a performance measure, the classification accuracy of dynamic pixels is chosen. 
Ideally, all dynamic parts of an image are masked out, yet the approach must not be over-conservative by masking out static parts of an image. 
In the following, different versions of our presented approach will be compared to the baseline solution, the semantic segmentation by the Mask \mbox{R-CNN} model~\cite{he2017mask}.

\subsection{Model training}
\paragraph{Training data}
For training the model, RGB images and the corresponding dynamic instance masks need to be provided.
We use a combination of synthetic and real-world data to train our model. 
The synthetic data is generated by the CARLA~\cite{dosovitskiy2017carla} simulator, which also allows for the extraction of semantic labels and therefore typical dynamic classes like pedestrians, cars or motorbikes. 

Only using this synthetic and semantically classified data would restrict our approach to a pre-defined set of semantic classes.
Therefore, we also recorded our own real-world dataset in Zurich.
This also allows to find general dynamic objects which might not be part of the synthetic data, \emph{e.g.}, parts of trees, animals or even water. 
The dynamic instance masks are extracted from the real-world data using the approach presented in Section~\ref{sec:approach}.
For both datasets, we covered different weather conditions and times of the day. 



\paragraph{Training procedure}
During training we always use a pre-trained version of the \emph{backbone} part to avoid learning feature extraction from scratch. 
The top layers (model \emph{head}) are initialized randomly due to the change in the output classes. 

The training phase is divided into two stages.
During the first training stage, only the model \emph{head} layers are trained while the pre-trained \emph{backbone} layers remain fixed. 
The training in stage 1 is conducted purely with synthetic data, and is run for 50 epochs. 
During the second training stage (50 epochs), three different models are trained: \modelmix with a combination of the real-world data and the synthetic data (50/50 ratio),  \modelreal with only real-world data, and \modelsynth with only synthetic data.
The learning rates are set as $10^{-3}$ and $10^{-4}$ in stages one and two, respectively.

\subsection{Evaluation}
In order to test the generalization of the presented approach to other cities and datasets, the quantitative evaluation will be conducted on the \emph{Cityscapes} dataset \cite{cordts2016cityscapes}.
From there, the \emph{Fine} subset was selected as it offers accurate semantic segmentation. 
All dynamic classes (except sky) are fused into a single \emph{dynamic} one which allows for the binary mask extraction of the ground truth data.
The performance indicator for the models is chosen to be the \fone score computed based on all pixel classifications between the ground truth and predicted binary image mask for each frame. 

\begin{figure}[htbp]
\centering
\adjincludegraphics[trim={{0.0\width} {0.92\height} {0.0\width} {0.01\height}}, clip, width=1.0\linewidth]{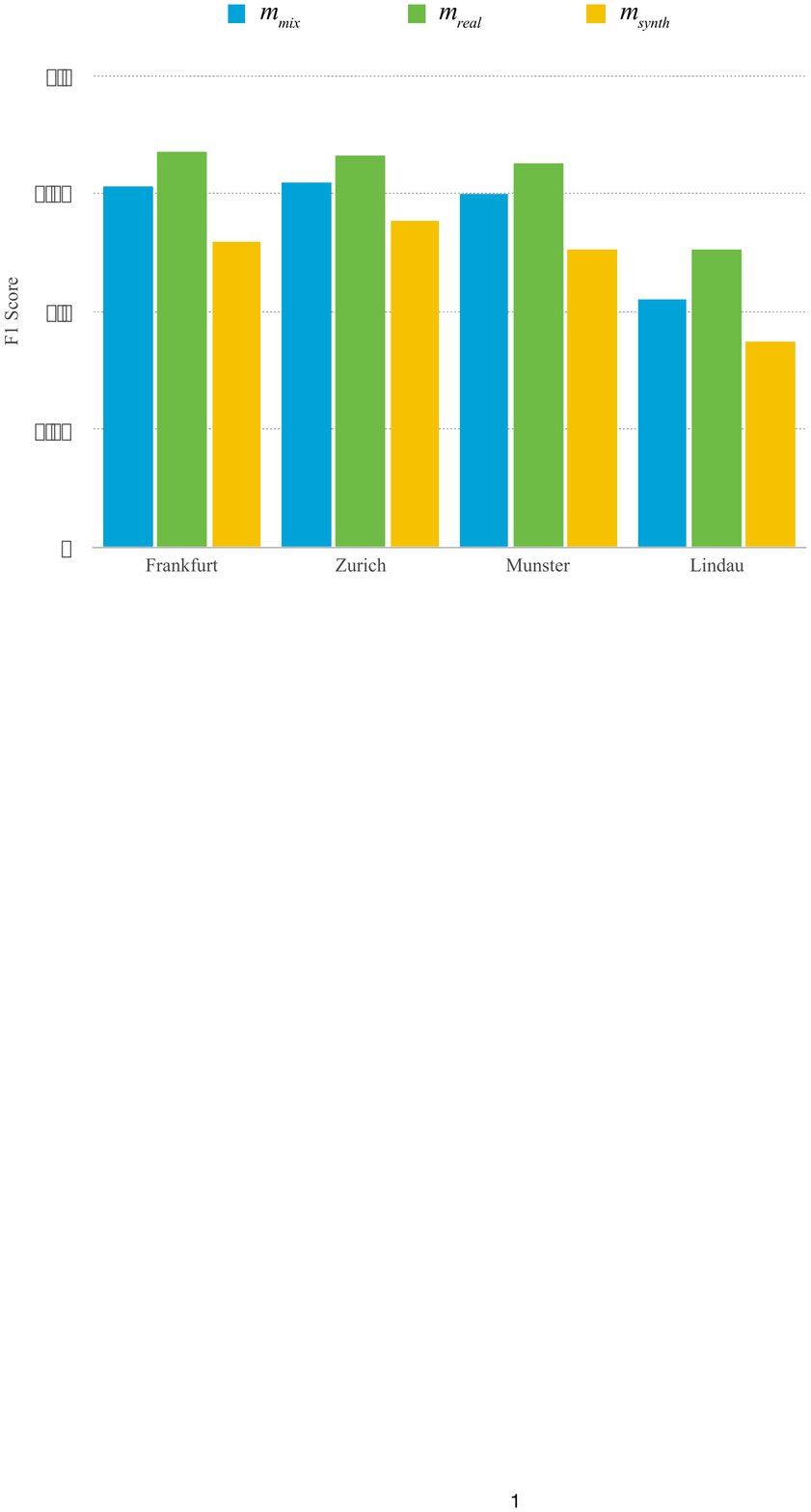}
\adjincludegraphics[trim={{0.0\width} {0.30\height} {0.0\width} {0.27\height}}, clip, width=1.0\linewidth]{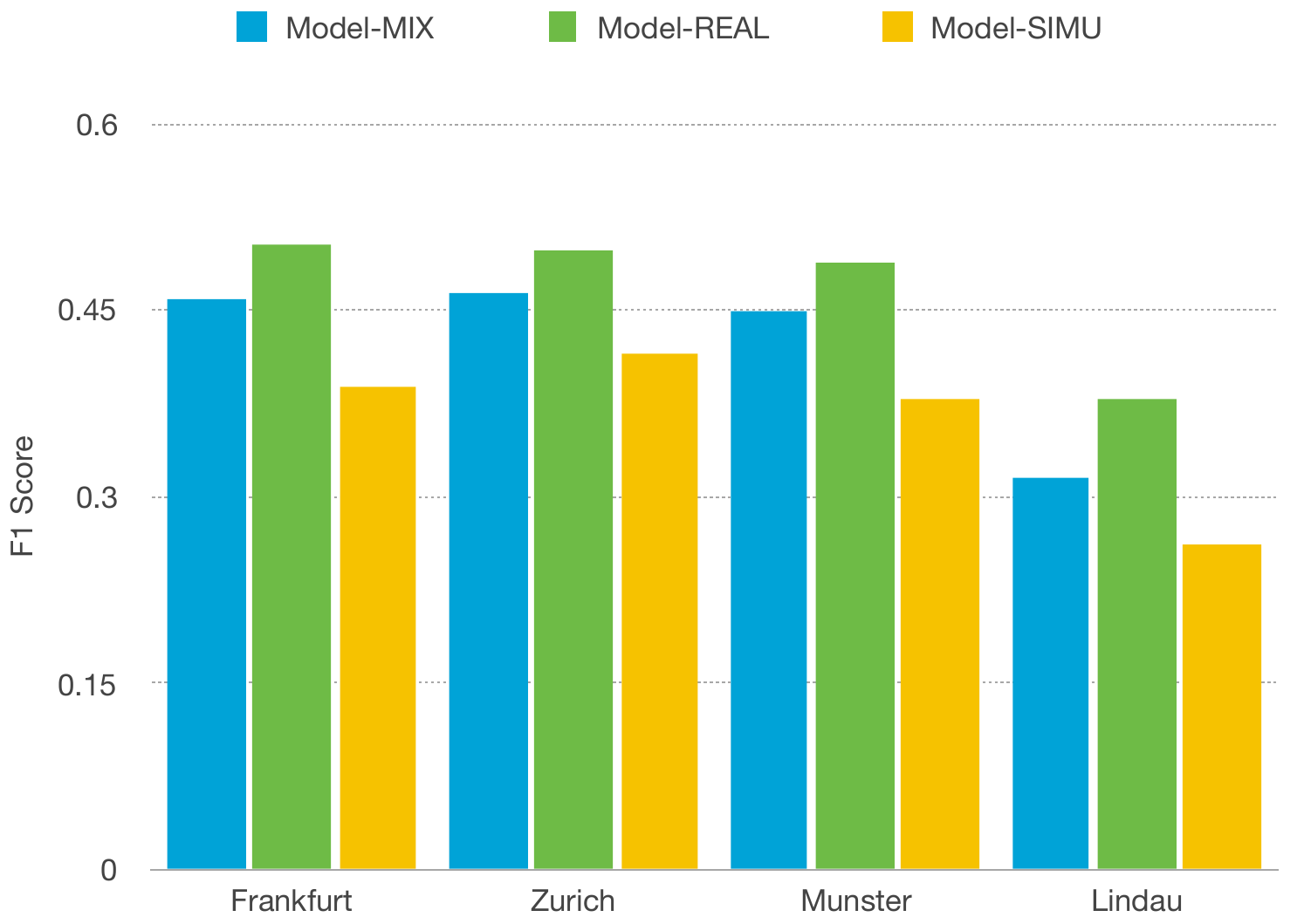}
\adjincludegraphics[trim={{0.0\width} {0.0\height} {0.0\width} {0.92\height}}, clip, width=1.0\linewidth]{chart.pdf}
\caption{\fone score of three models on different city image sets from \emph{Fine} Cityscapes dataset.
The model \modelreal ~ achieves the best performance in all locations, which proves the value of real-world data in the training set.
The masks to train the classifier were obtained using the methodology presented in Section~\ref{subsec:train_data}.}
    \label{fig:chart}
\end{figure}

Figure~\ref{fig:chart} shows the evaluation results of our three presented models in four city environments.
The results show that \modelreal ~ performs the best on the test set, while \modelsynth ~ reaches a significantly worse \fone score. 
Given that \modelsynth ~ has only seen real-world data during pre-training but in none of the two further training steps, it might be tailored to the rendered synthetic data and fail to extract proper features in real-world environments.
\modelreal ~ shows that after pre-training the backbone with simulation data, it is better to only refine the network on real-world data and not a mixture between the two. 

\begin{figure}
    \centering
    \includegraphics[width=\columnwidth]{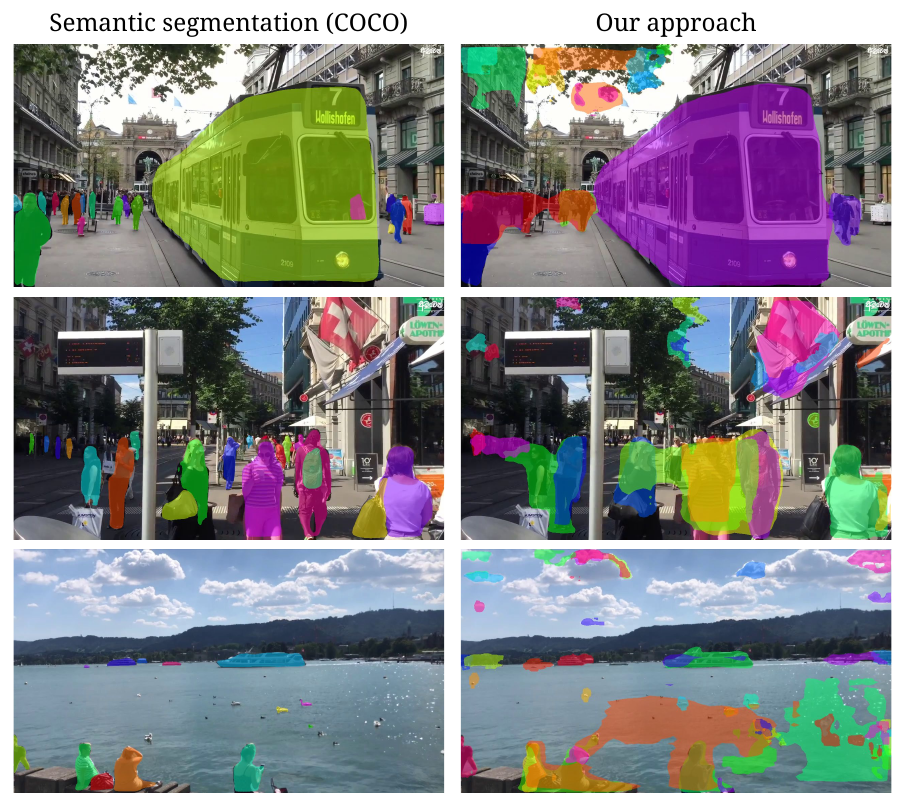}
    \caption{Qualitative analysis of the proposed dynamic instance detection compared to a standard semantic segmentation.
    Note how our algorithm trained on the real-world data manages to capture the location-specific dynamic objects, such as trees, flags or animals.}
    \label{fig:qual-eval}
    \vspace{-5mm}
\end{figure}

Figure~\ref{fig:qual-eval} shows various image frames of the detected dynamic object masks of our presented framework and their semantic segmentation.
The results of the proposed method are generated using \modelreal, which performs the best according to Figure~\ref{fig:chart}.
The semantic segmentation is based on~\cite{he2017mask}.
The qualitative results demonstrate superior performance of a classifier trained using real-world data that manages to capture even the location-specific objects.

Furthermore, we provide a video\footnote{\href{https://www.youtube.com/watch?v=CnqrSFfhpxM}{\tt https://www.youtube.com/watch?v=CnqrSFfhpxM}} where our best model (\modelreal) is applied to image frames recorded in city conditions.

\section{Conclusion}
\label{sec:conclusion}
In this paper, we presented an approach to detect dynamic object instances to improve the robustness of vision-based \ac{slam} in dynamic environments. 
We propose to use a combination of easily accessible simulation data and more expensive but specific real-world data to train a \ac{cnn} that generalizes well to unseen environments, such as Cityscapes.
Additionally, we introduce an approach to extract the dynamic object masks from our real-world dataset in an automated fashion.
We also demonstrate that real-world data is crucial to outperform models purely trained with synthetic data.

While in many applications a pure semantic analysis might be enough to segment out dynamic objects, especially, when moving towards unknown environments a more general approach like ours becomes valuable. 
General dynamic objects can be detected, even if no prior segmentation classes are provided. 

Future work will fully integrate this approach into a vision-based localization system for dynamic environments.

\footnotesize
\bibliographystyle{IEEEtran}
\balance
\bibliography{bib/IEEEfull.bib,bib/mark_references}

\begin{thebibliography}{10}
\providecommand{\url}[1]{#1}
\csname url@samestyle\endcsname
\providecommand{\newblock}{\relax}
\providecommand{\bibinfo}[2]{#2}
\providecommand{\BIBentrySTDinterwordspacing}{\spaceskip=0pt\relax}
\providecommand{\BIBentryALTinterwordstretchfactor}{4}
\providecommand{\BIBentryALTinterwordspacing}{\spaceskip=\fontdimen2\font plus
\BIBentryALTinterwordstretchfactor\fontdimen3\font minus
  \fontdimen4\font\relax}
\providecommand{\BIBforeignlanguage}[2]{{%
\expandafter\ifx\csname l@#1\endcsname\relax
\typeout{** WARNING: IEEEtran.bst: No hyphenation pattern has been}%
\typeout{** loaded for the language `#1'. Using the pattern for}%
\typeout{** the default language instead.}%
\else
\language=\csname l@#1\endcsname
\fi
#2}}
\providecommand{\BIBdecl}{\relax}
\BIBdecl

\bibitem{pfeiffer2016iros}
M.~Pfeiffer, U.~Schwesinger \emph{et~al.}, ``Predicting actions to act
  predictably: Cooperative partial motion planning with maximum entropy
  models,'' in \emph{{Proc. of IEEE/RSJ Int. Conf. on Intelligent Robots and
  Systems (IROS)}}.\hskip 1em plus 0.5em minus 0.4em\relax IEEE, Oct. 2016, pp.
  2096--2101.

\bibitem{pfeiffer2017lstm}
M.~Pfeiffer, G.~Paolo \emph{et~al.}, ``A data-driven model for
  interaction-aware pedestrian motion prediction in object cluttered
  environments,'' \emph{arXiv preprint arXiv:1709.08528}, 2017.

\bibitem{kretzschmar2016social}
H.~Kretzschmar, M.~Spies \emph{et~al.}, ``Socially compliant mobile robot
  navigation via inverse reinforcement learning,'' \emph{The International
  Journal of Robotics Research}, vol.~35, no.~11, pp. 1289--1307, 2016.

\bibitem{vemula2017modeling}
A.~Vemula, K.~Muelling, and J.~Oh, ``Modeling cooperative navigation in dense
  human crowds,'' \emph{arXiv preprint arXiv:1705.06201}, 2017.

\bibitem{pettre2009experiment}
J.~Pettr{\'e}, J.~Ond{\v{r}}ej \emph{et~al.}, ``Experiment-based modeling,
  simulation and validation of interactions between virtual walkers,'' in
  \emph{Proceedings of the 2009 ACM SIGGRAPH/Eurographics Symposium on Computer
  Animation}.\hskip 1em plus 0.5em minus 0.4em\relax ACM, 2009, pp. 189--198.

\bibitem{schneider2018maplab}
T.~Schneider, M.~Dymczyk \emph{et~al.}, ``maplab: An open framework for
  research in visual-inertial mapping and localization,'' \emph{IEEE Robotics
  and Automation Letters}, vol.~3, no.~3, pp. 1418--1425, 2018.

\bibitem{lynen2015get}
S.~Lynen, T.~Sattler \emph{et~al.}, ``Get out of my lab: Large-scale, real-time
  visual-inertial localization.'' in \emph{RSS}, 2015.

\bibitem{mur2015orb}
R.~Mur-Artal, J.~M.~M. Montiel, and J.~D. Tardos, ``{ORB-SLAM: a versatile and
  accurate monocular SLAM system},'' \emph{IEEE Transactions on Robotics},
  vol.~31, no.~5, pp. 1147--1163, 2015.

\bibitem{he2017mask}
K.~He, G.~Gkioxari \emph{et~al.}, ``Mask r-cnn,'' in \emph{Computer Vision
  (ICCV), 2017 IEEE International Conference on}.\hskip 1em plus 0.5em minus
  0.4em\relax IEEE, 2017, pp. 2980--2988.

\bibitem{dosovitskiy2017carla}
A.~Dosovitskiy, G.~Ros \emph{et~al.}, ``Carla: An open urban driving
  simulator,'' \emph{arXiv preprint arXiv:1711.03938}, 2017.

\bibitem{klein2007parallel}
G.~Klein and D.~Murray, ``{Parallel tracking and mapping for small AR
  workspaces},'' in \emph{Mixed and Augmented Reality, 2007. ISMAR 2007. 6th
  IEEE and ACM International Symposium on}, 2007, pp. 225--234.

\bibitem{alcantarilla2012combining}
P.~F. Alcantarilla, J.~J. Yebes \emph{et~al.}, ``{On combining visual SLAM and
  dense scene flow to increase the robustness of localization and mapping in
  dynamic environments},'' in \emph{Robotics and Automation (ICRA), 2012 IEEE
  International Conference on}.\hskip 1em plus 0.5em minus 0.4em\relax IEEE,
  2012, pp. 1290--1297.

\bibitem{tan2013robust}
W.~Tan, H.~Liu \emph{et~al.}, ``{Robust monocular SLAM in dynamic
  environments},'' in \emph{Mixed and Augmented Reality (ISMAR), 2013 IEEE
  International Symposium on}, 2013, pp. 209--218.

\bibitem{kim2016effective}
D.-H. Kim and J.-H. Kim, ``{Effective Background Model-Based RGB-D Dense Visual
  Odometry in a Dynamic Environment},'' \emph{IEEE Transactions on Robotics},
  vol.~32, no.~6, pp. 1565--1573, 2016.

\bibitem{li2017rgb}
S.~Li and D.~Lee, ``{RGB-D SLAM in Dynamic Environments Using Static Point
  Weighting},'' \emph{IEEE Robotics and Automation Letters}, vol.~2, no.~4, pp.
  2263--2270, 2017.

\bibitem{sun2017improving}
Y.~Sun, M.~Liu, and M.~Q.-H. Meng, ``{Improving RGB-D SLAM in dynamic
  environments: A motion removal approach},'' \emph{Robotics and Autonomous
  Systems}, vol.~89, pp. 110--122, 2017.

\bibitem{badrinarayanan2015segnet}
V.~Badrinarayanan, A.~Kendall, and R.~Cipolla, ``Segnet: A deep convolutional
  encoder-decoder architecture for image segmentation,'' \emph{arXiv preprint
  arXiv:1511.00561}, 2015.

\bibitem{guizilini2015online}
V.~Guizilini and F.~Ramos, ``Online self-supervised learning for dynamic object
  segmentation,'' \emph{The International Journal of Robotics Research},
  vol.~34, no. 4-5, pp. 559--581, 2015.

\bibitem{barnes2017driven}
D.~Barnes, W.~Maddern \emph{et~al.}, ``Driven to distraction: Self-supervised
  distractor learning for robust monocular visual odometry in urban
  environments,'' \emph{arXiv preprint arXiv:1711.06623}, 2017.

\bibitem{hartmann2014predicting}
W.~Hartmann, M.~Havlena, and K.~Schindler, ``Predicting matchability,'' in
  \emph{Proceedings of the IEEE Conference on Computer Vision and Pattern
  Recognition}, 2014, pp. 9--16.

\bibitem{dymczyk2016will}
M.~Dymczyk, E.~Stumm \emph{et~al.}, ``Will it last? learning stable features
  for long-term visual localization,'' in \emph{3D Vision (3DV), 2016 Fourth
  International Conference on}.\hskip 1em plus 0.5em minus 0.4em\relax IEEE,
  2016, pp. 572--581.

\bibitem{bescos2018dynslam}
B.~Besc{\'o}s, J.~M. F{\'a}cil \emph{et~al.}, ``Dynslam: Tracking, mapping and
  inpainting in dynamic scenes,'' \emph{arXiv preprint arXiv:1806.05620}, 2018.

\bibitem{girshick2015fast}
R.~Girshick, ``Fast r-cnn,'' in \emph{Proceedings of the IEEE international
  conference on computer vision}, 2015, pp. 1440--1448.

\bibitem{lin2014microsoft}
T.-Y. Lin, M.~Maire \emph{et~al.}, ``Microsoft coco: Common objects in
  context,'' in \emph{European conference on computer vision}.\hskip 1em plus
  0.5em minus 0.4em\relax Springer, 2014, pp. 740--755.

\bibitem{ren2003learning}
X.~Ren and J.~Malik, ``Learning a classification model for segmentation,'' in
  \emph{null}.\hskip 1em plus 0.5em minus 0.4em\relax IEEE, 2003, p.~10.

\bibitem{cordts2016cityscapes}
M.~Cordts, M.~Omran \emph{et~al.}, ``The cityscapes dataset for semantic urban
  scene understanding,'' in \emph{Proc. of the IEEE Conference on Computer
  Vision and Pattern Recognition (CVPR)}, 2016.

\end{thebibliography}

\end{document}